\documentclass[letterpaper]{article} 
\usepackage{aaai25}  
\usepackage{times}  
\usepackage{helvet}  
\usepackage{courier}  
\usepackage[hyphens]{url}  
\usepackage{graphicx} 
\usepackage{natbib}  
\usepackage{caption} 
\usepackage{algorithm}
\usepackage{algorithmic}

\usepackage{enumitem}
\usepackage{romannum}
\usepackage{booktabs}
\usepackage{amsmath}
\usepackage{cuted}
\usepackage{capt-of}

\usepackage{array}
\newcolumntype{P}[1]{>{\centering\arraybackslash}p{#1}}
\usepackage{amssymb}
\usepackage{multirow}

\usepackage{newfloat}
\usepackage{listings}
\DeclareCaptionStyle{ruled}{labelfont=normalfont,labelsep=colon,strut=off} 
\floatstyle{ruled}
\newfloat{listing}{tb}{lst}{}
\floatname{listing}{Listing}

\nocopyright 

\title{Improving Unsupervised Video Object Segmentation via Fake Flow Generation}

\author{
Suhwan Cho\quad Minhyeok Lee\quad Jungho Lee\quad Donghyeong Kim\\
Seunghoon Lee\quad Sungmin Woo\quad Sangyoun Lee
}

\affiliations{
Yonsei University, Seoul, Korea\\
chosuhwan@yonsei.ac.kr\quad suhwanx@gmail.com
}

\begin{document}
\maketitle

\begin{abstract}
Unsupervised video object segmentation (VOS), also known as video salient object detection, aims to detect the most prominent object in a video at the pixel level. Recently, two-stream approaches that leverage both RGB images and optical flow maps have gained significant attention. However, the limited amount of training data remains a substantial challenge. In this study, we propose a novel data generation method that simulates fake optical flows from single images, thereby creating large-scale training data for stable network learning. Inspired by the observation that optical flow maps are highly dependent on depth maps, we generate fake optical flows by refining and augmenting the estimated depth maps of each image. By incorporating our simulated image-flow pairs, we achieve new state-of-the-art performance on all public benchmark datasets without relying on complex modules. We believe that our data generation method represents a potential breakthrough for future VOS research.
Code and models are available at \url{https://github.com/suhwan-cho/FakeFlow}.
\end{abstract}

\begin{figure}[t]
\centering
\includegraphics[width=1\linewidth]{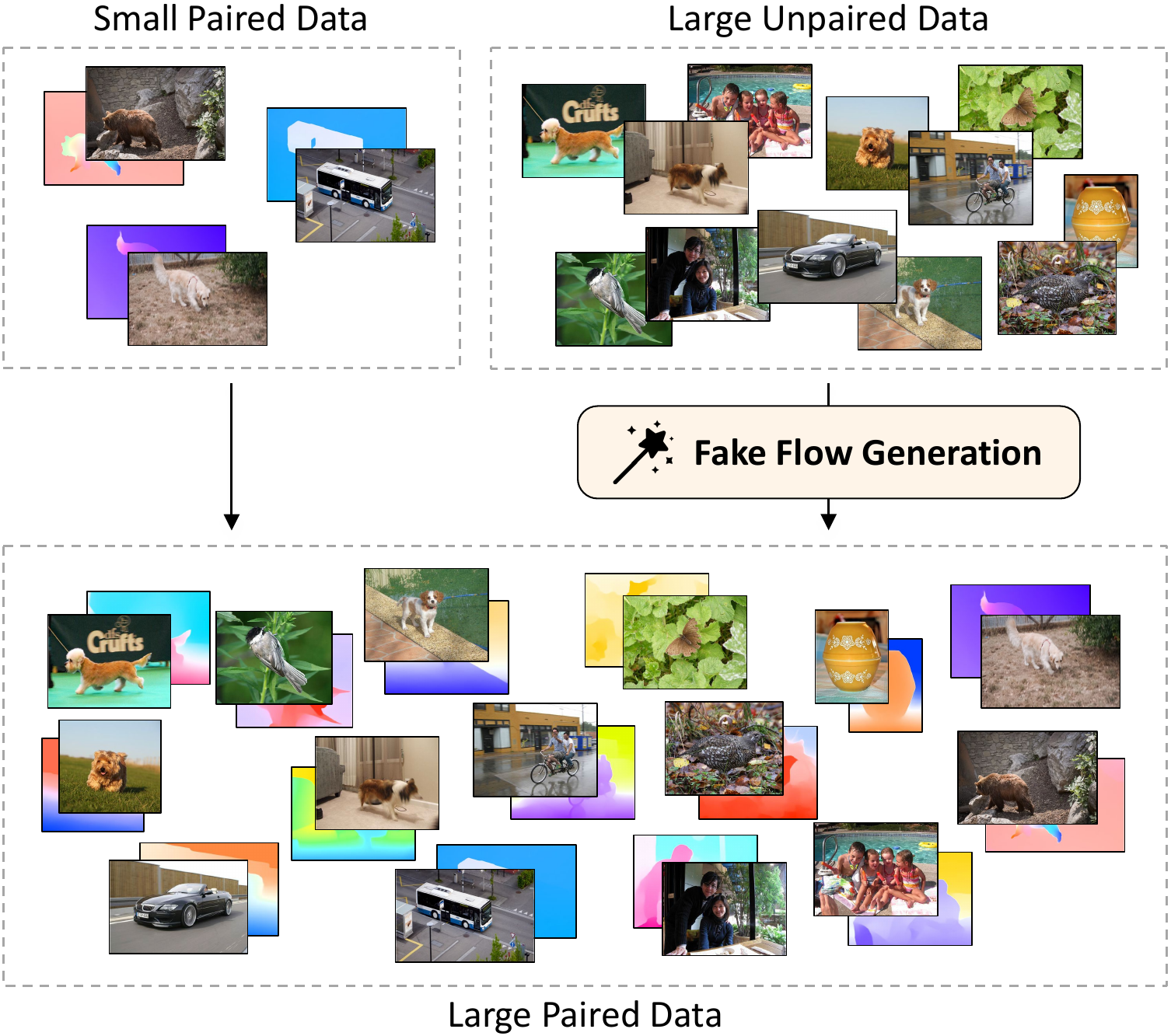}
\caption{Visualization of our proposed training data construction strategy. By simulating optical flow maps from a large number of single images, large-scale image-flow pairs can be used for network training.}
\label{figure1}
\end{figure}

\section{Introduction}
Video object segmentation (VOS) is a fundamental task in computer vision, serving as an essential component for video processing. Depending on the use cases, VOS can incorporate external guidance such as initial frame masks, initial frame points, human interaction, and language guidance. In this study, we focus on the unsupervised setting, which aims to segment the most salient object in a video without any external guidance. From the input video sequence, we must first define the most prominent object and then identify that object at the pixel level.

In unsupervised VOS, also known as video salient object detection (SOD), two-stream approaches have recently gained significant attention. Optical flow maps between adjacent frames are first extracted from RGB frames using an optical flow estimation model. For each frame, an RGB image and the corresponding flow map are then used as source information to produce the segmentation mask. However, the scarcity of video-level SOD data has always been a bottleneck. To address this, existing methods leverage large-scale VOS data or image-level SOD data for network training. However, the former may include non-salient objects and the latter does not contain optical flow maps, preventing complete network training. This necessitates an additional fine-tuning process on the small-scale video-level SOD data, which is insufficient for training a robust network.

Inspired by two key observations on two-stream VOS approaches, we address the scarcity of training data. First, most two-stream methods do not utilize long-term coherence reasoning. Since salient objects typically exhibit distinctive movements compared to the background, these methods generally operate on a per-frame basis, leveraging only the short-term properties captured by optical flow maps. This indicates that what we actually need are image-flow pairs rather than long video sequences. Second, the precise pixel values in each flow map are not crucial. In two-stream methods, the primary role of optical flow maps is to help distinguish foreground objects from the background. In other words, we aim to capture the distinctive movements of the objects themselves, not the exact movement values for warping. Therefore, in each image-flow pair, the flow maps do not need to be accurate; plausible flow maps are sufficient.

Based on the aforementioned observations, we propose a novel data generation method that simulates optical flow maps from single images, as depicted in Figure~\ref{figure1}. Our method is founded on the simple assumption that optical flow maps are highly dependent on depth maps, which can be easily derived. For instance, moving objects that exhibit distinctive movements compared to the background generally have distinctive depths as well. Furthermore, the distinctive movements of static objects are caused by depth differences, aligning with our assumption. We further argue that our fake flow simulation may not follow real optical flows due to vagueness in its nature, but this serves as challenging training data during network training.

Our proposed fake flow simulation is straightforward. From each RGB image, we estimate the depth map using a pre-trained depth estimation model. Then, we normalize the depth maps to have values between 0 and 1, with the values randomly reversed for data diversity. The normalized depth maps are then duplicated into two motion maps, indicating motion along the $x$-axis and the $y$-axis, respectively. Each motion map is then re-scaled from -1 to 1 to simulate positive and negative motion values. Finally, we randomly apply value shifting and scaling for data augmentation. The 2-channel UV maps are converted to 3-channel RGB flow maps and serve as input information.

Following common two-stream methods \cite{HFAN, SimulFlow, DPA}, we design our network based on a simple encoder-decoder architecture. Without any complex modules or post-processing steps, our model trained with fake optical flows sets a new state-of-the-art performance on all public benchmark datasets. Our flow simulation method also significantly improves existing VOS methods, demonstrating its scalability. We believe our method can serve as a powerful tool to remove the common bottleneck of unsupervised VOS.

Our main contributions can be summarized as follows:
\begin{itemize}[leftmargin=0.2in]
\item We introduce a novel data generation method to leverage large-scale image-flow pairs during network training by simulating optical flow maps from single images.

\item Using our simulated data, we set a new state-of-the-art performance on all public benchmark datasets with a simple encoder-decoder architecture.

\item We construct a new dataset comprising image-flow pairs, facilitating future research in VOS and related fields.
\end{itemize}

\begin{figure*}[t]
\centering
\includegraphics[width=1\linewidth]{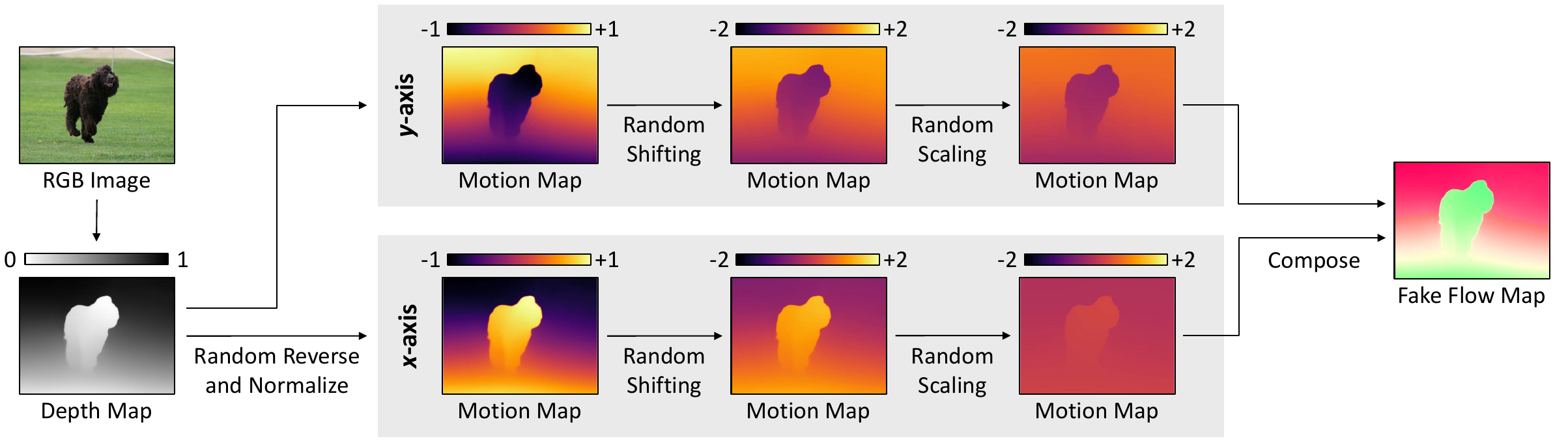}
\caption{Overall pipeline of our fake flow generation process.}
\label{figure2}
\end{figure*}

\section{Related Work}
\noindent\textbf{Temporal reasoning.}
In video sequences, frames often share similar properties, even when temporally distant, creating opportunities for leveraging temporal coherence as a powerful cue for unsupervised video object segmentation (VOS). Several methods have been proposed to utilize this property. AGS~\cite{AGS} employs recurrent neural networks to maintain spatial details while simultaneously modeling temporal dynamics. COSNet~\cite{COSNet} introduces a co-attention module that enables UVOS by densely comparing pairs of frames using an attention mechanism, capturing rich correlations at every pixel location. AGNN~\cite{AGNN} employs attentive graph neural networks for dense feature propagation between frames, treating each frame as a node and the relationships between frames as edges. DFNet~\cite{DFNet} presents a discriminative feature learning network that models long-term correlations by representing the entire feature distribution of the input video. AD-Net~\cite{AD-Net} utilizes the initial frame as an anchor to learn pairwise dependencies for each query frame prediction, while F2Net~\cite{F2Net} enhances AD-Net with point heatmap estimation to provide spatial priors during mask prediction. Additionally, 3DC-Seg~\cite{3DC-Seg} and D$^2$Conv3D~\cite{D2Conv3D} use 3D convolution architectures to explicitly consider both spatial and temporal connectivity. IMP~\cite{IMP} adopts a unique approach by sequentially connecting image-level mask predictions with the temporal propagation of these predictions.

\vspace{1mm}
\noindent\textbf{Multi-modality fusion.}
Recognizing that RGB images can sometimes be ambiguous, several methods have implemented two-stream architectures that integrate motion cues from optical flow maps to supplement appearance cues. The main objective of these two-stream approaches is to effectively fuse different modalities. MATNet~\cite{MATNet} utilizes deeply interleaved encoders to facilitate hierarchical interactions between object appearance and motion cues. RTNet~\cite{RTNet} introduces a reciprocal transformation network that merges appearance and motion features. AMC-Net~\cite{AMC-Net} employs a co-attention gate to balance contributions from multi-modal features and suppress misleading information through explicit scoring. TransportNet~\cite{TransportNet} establishes correspondences between modalities while mitigating distracting signals using optimal structural matching. FSNet~\cite{FSNet} proposes a mutual feature propagation network to provide mutual restraint within a full-duplex strategy. HFAN~\cite{HFAN} features a hierarchical architecture that aligns and fuses appearance and motion features across multiple embedding layers, while SimulFlow~\cite{SimulFlow} conducts feature extraction and target identification at the joint embedding stage of both appearance and motion.

Some methods based on the two-stream architecture incorporate additional techniques to enhance robustness. TMO~\cite{TMO} trains a flow encoder that combines both appearance and motion cues to improve resilience against low-quality optical flows. OAST~\cite{OAST} and DATTT~\cite{DATTT} use test-time learning protocols to refine the trained network during inference. GFA~\cite{GFA} augments training samples to align data distributions between the training and testing phases.

\vspace{1mm}
\noindent\textbf{Hybrid structure.}
Recent approaches leverage both temporal coherence for long-term consistency and motion information for robust short-term cues. PMN~\cite{PMN} employs an external memory bank to store and continuously update appearance and motion features, enhancing the online representation of object properties. GSA-Net~\cite{GSA-Net} and DPA~\cite{DPA} utilize a global property injection module with reference frames, providing the network with comprehensive knowledge of each video sequence.

\section{Approach}

\subsection{Preliminaries}

\noindent\textbf{Conventional pipeline.}
Considering the power of optical flow maps for detecting primary objects, recent unsupervised VOS methods are mostly based on the two-stream architecture, i.e., RGB images $I$ and optical flow maps $F$ are given as source information for feature extraction. Formally, optical flows are calculated as
\begin{align}
&F^i = \Phi(I^i, I^{i+1}),
\end{align}
where $\Phi(I_1,I_2)$ indicates optical flow from the source frame $I_1$ to the target frame $I_2$, and $i$ denotes the video frame index. Note that for the last frame, the target frame is the frame before the source frame. With $I$ and $F$ of each frame, the segmentation mask $M$ is obtained as
\begin{align}
&M = \Psi(I, F; \mathrm{w}),
\end{align}
where $\Psi$ and $\mathrm{w}$ denote the network to be learned and its learnable weights, respectively. Generally, this process is implemented in a per-frame manner, assuming the short-term temporal cues can provide enough information to detect the salient object in videos.

\vspace{1mm}
\noindent\textbf{Training protocol.}
Although the two-stream VOS architecture is powerful, training a generalized and robust network is quite challenging due to the limited availability of video-level SOD data. This scarcity has always been a bottleneck in the field. Specifically, most existing methods use the DAVIS 2016~\cite{DAVIS} training set as their primary dataset. However, this set only contains 30 video sequences, which provides insufficient data diversity for effective generalized network learning.

To address this issue, two approaches have been widely adopted: using large-scale VOS datasets and image-level SOD datasets. Some methods~\cite{HFAN, SimulFlow, GSA-Net} utilize large-scale VOS datasets such as YouTube-VOS 2018~\cite{YTVOS}. Since the video samples in this dataset contain multiple annotated objects, the multi-label masks are converted to binary masks by defining all objects as salient. However, these objects are often not actually salient, which introduces noise during network training. Consequently, the YouTube-VOS dataset is typically used as a pre-training dataset in two-stream approaches. Alternatively, some methods~\cite{FSNet, PMN, TMO} leverage image-level SOD data such as DUTS~\cite{DUTS} for network training. Due to the unavailability of optical flow maps in these samples, these methods partially train the network with the motion branch frozen. The motion branch is then trained at a later stage using video-level SOD data.

While these methods help mitigate the scarcity of training data to some extent, there still remains a disparity between the properties of these datasets and video-level SOD datasets. This necessitates an additional fine-tuning process on the small-scale video-level SOD data, which can lead to limitations such as network overfitting.

\begin{figure}[t]
\centering
\includegraphics[width=1\linewidth]{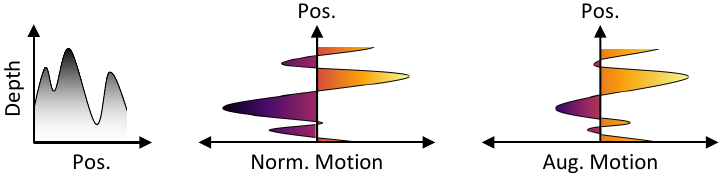}
\caption{Visualized depth-to-flow conversion process. Augmented motion indicates the motion map after applying random value shifting and scaling.}
\label{figure3}
\end{figure}

\begin{figure*}[t!]
\centering
\includegraphics[width=1\linewidth]{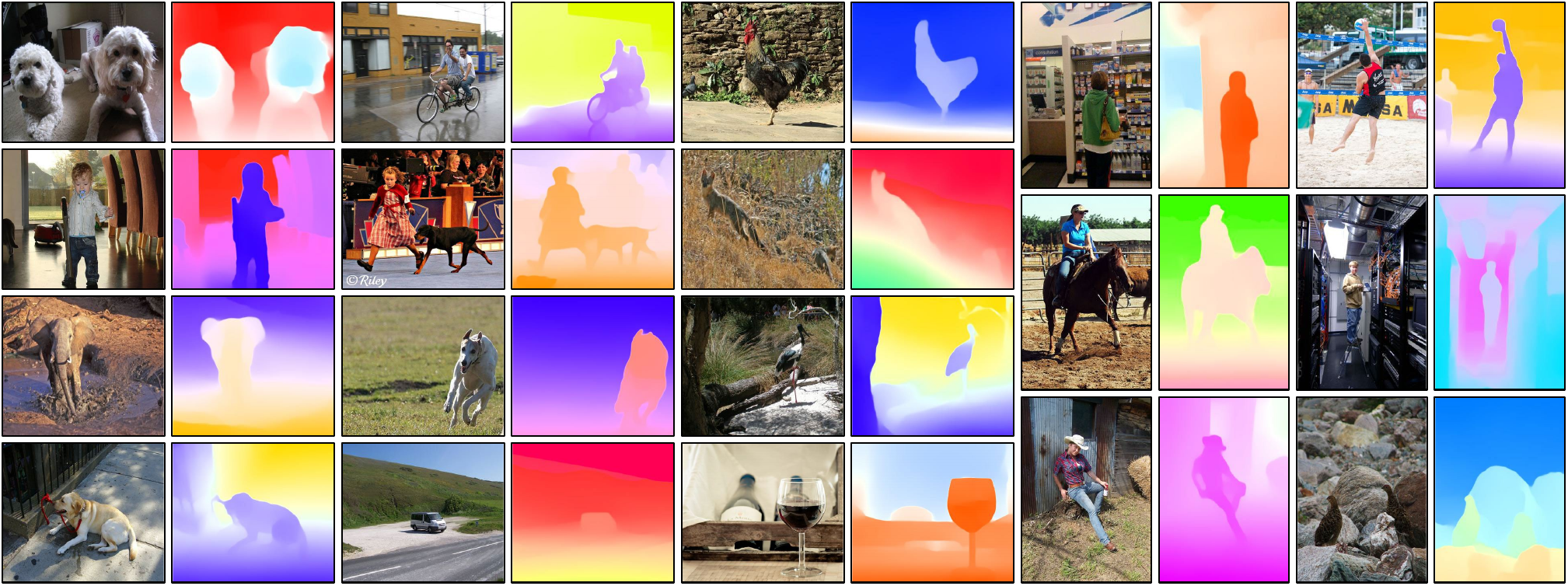}
\caption{Example image-flow pairs from DUTSv2 dataset.}
\label{figure4}
\end{figure*}

\vspace{1mm}
\noindent\textbf{Key observations.}
Our study starts from two key observations on the existing two-stream VOS approaches. First, long-term coherence reasoning is typically not required. The short-term cues from frame-level flow maps are usually sufficient for capturing saliency in videos, leading most existing methods to operate in a per-frame manner. This implies that what we need are image-flow pairs rather than long video sequences. Second, precise pixel values are not crucial when leveraging optical flow maps. What we aim to obtain are the properties that can separate the foreground from the background. Therefore, the flow maps should be plausible rather than accurate; perceptually high-quality flows are more useful than flows with low error rates in actual movements.

\subsection{Fake Flow Generation}
Considering the aforementioned characteristics and trends in unsupervised VOS research, we propose a novel method to overcome the scarcity of video-level SOD training data. Since optical flow maps are often closely related to depth maps, we simulate fake optical flows from the estimated depth maps. Figure~\ref{figure2} visualizes the overall pipeline of our fake flow generation process.

\vspace{1mm}
\noindent\textbf{Depth estimation.}
To simulate fake optical flows from single images, we first estimate the depth of each image using a pre-trained depth estimation model, specifically, the DPT-Hybrid~\cite{DPT} model. Given that the depth estimation model is denoted as $\Omega$, the normalized depth map $D \in [0,1]$ can be obtained as
\begin{align}
&D = \frac{\Omega(I) - \mathrm{Min} \big(\Omega(I)_p \big)}{\mathrm{Max} \big(\Omega(I)_p \big) - \mathrm{Min} \big(\Omega(I)_p \big)}~,
\end{align}
where $p$ indicates each pixel location.

\vspace{1mm}
\noindent\textbf{Depth-to-flow conversion.}
With the normalized depth map $D$, our objective is to convert it to a simulated flow map. This requires converting the depth map to two channels and considering both positive and negative directions. Additionally, we introduce some randomness to enhance data diversity. Note that the depth-to-flow conversion is applied independently to both the $x$-axis and $y$-axis.

The first step involves transforming the depth map with a random value reverse for data diversity and normalization as
\begin{align}
&M' = 2 \delta (1-D) + 2 (1-\delta) D - 1, \quad \delta \overset{\mathrm{R}}{\leftarrow} \{0, 1\}~.
\end{align}
After obtaining the motion map $M'$, we apply random augmentation techniques to create diverse scenarios for network training. The first augmentation is random value shifting, which can be denoted as
\begin{align}
&M'' = M' + \epsilon, \quad \epsilon \overset{\mathrm{R}}{\leftarrow} [-1, 1]~.
\end{align}
By adding a random value to every pixel location, we effectively simulate both positive and negative pixel movements at the same time, enhancing motion diversity. Next, we consider the absolute value scale of each axis to simulate various cases for optical flow diversity. We apply random scaling to each motion map as
\begin{align}
&M''' = \eta M'', \quad \eta \overset{\mathrm{R}}{\leftarrow} [0, 1]~.
\end{align}
In Figure~\ref{figure3}, we visualize the depth-to-flow conversion process. Starting from the depth map, we generate a normalized motion map. By applying augmentation techniques, we simulate significantly diverse scenarios.

\vspace{1mm}
\noindent\textbf{Flow visualization.}
After the depth-to-flow conversion, we have a two-channel augmented motion map $M''' \in [-2, 2]$. Now, we transform the motion map into an RGB optical flow map to use it with conventional feature extractors. The UV-to-RGB conversion is formally denoted as
\begin{align}
&F = \Upsilon \Big(\frac{M'''}{\mathrm{Max}(\lVert{M'''_p}\rVert_2)} \Big)~,
\end{align}
where $\Upsilon$ indicates a pre-defined mapping function that transits a 2-channel value to a 3-channel value.

\vspace{1mm}
\noindent\textbf{Data processing.}
We use the fake flow generation protocol to construct large-scale training data for the two-stream VOS architecture. To achieve this, we choose a large-scale image-level SOD dataset, DUTS, as our source dataset. We collect both training and testing images from the dataset and generate fake flow maps for all the image samples. The constructed image-flow pairs are used as training data for our method during the main network training stage, and we refer to our DUTS variant as DUTSv2. In Figure~\ref{figure4}, we visualize some example image-flow pairs in our DUTSv2 dataset.

For the video data used for both training and testing, the optical flow maps are generated using RAFT~\cite{RAFT} while maintaining the original resolution of each video sample. In video datasets, each video frame containing RGB images and optical flow maps is considered as an image-flow pair in DUTSv2.

\begin{table*}[t]
\centering 
\small
\begin{tabular}{p{2cm}P{2cm}P{2.1cm}P{2cm}P{6mm}P{6mm}P{6mm}P{8mm}P{8mm}P{8mm}P{8mm}}
\toprule
\multicolumn{7}{c}{} &\multicolumn{3}{c}{DAVIS 2016} &\multicolumn{1}{c}{FBMS}\\
\cmidrule(lr){8-11}
Method &Publication &Backbone &Resolution &OF &PP &fps &$\mathcal{G}_\mathcal{M}$ &$\mathcal{J}_\mathcal{M}$ &$\mathcal{F}_\mathcal{M}$ &$\mathcal{J}_\mathcal{M}$\\
\midrule
MATNet &AAAI'20 &ResNet-101 &473$\times$473 &\checkmark &\checkmark &20.0 &81.6 &82.4 &80.7 &76.1\\
DFNet &ECCV'20 &DeepLabv3 &- & &\checkmark &3.57 &82.6 &83.4 &81.8 &-\\
F2Net &AAAI'21 &DeepLabv3 &473$\times$473 & & &10.0 &83.7 &83.1 &84.4 &77.5\\
RTNet &CVPR'21 &ResNet-101 &384$\times$672 &\checkmark &\checkmark &- &85.2 &85.6 &84.7 &-\\
FSNet &ICCV'21 &ResNet-50 &352$\times$352 &\checkmark &\checkmark &12.5 &83.3 &83.4 &83.1 &-\\
TransportNet &ICCV'21 &ResNet-101 &512$\times$512 &\checkmark & &12.5 &84.8 &84.5 &85.0 &78.7\\
AMC-Net &ICCV'21 &ResNet-101 &384$\times$384 &\checkmark &\checkmark &17.5 &84.6 &84.5 &84.6 &76.5\\
D$^2$Conv3D &WACV'22 &ir-CSN-152 &480$\times$854 & & &- &86.0 &85.5 &86.5 &-\\
IMP &AAAI'22 &ResNet-50 &- & & &1.79 &85.6 &84.5 &86.7 &77.5\\
HFAN &ECCV'22 &MiT-b2 &512$\times$512 &\checkmark & &12.8$^\ast$ &87.5 &86.8 &88.2 &-\\
PMN &WACV'23 &VGG-16 &352$\times$352 &\checkmark & &\underline{41.3}$^\ast$ &85.9 &85.4 &86.4 &77.7\\
TMO &WACV'23 &ResNet-101 &384$\times$384 &\checkmark & &\textbf{43.2}$^\ast$ &86.1 &85.6 &86.6 &79.9\\
OAST &ICCV'23 &MobileViT3D &384$\times$640 &\checkmark & &- &87.0 &86.6 &87.4 &83.0\\
SimulFlow &ACMMM'23 &MiT-b2 &512$\times$512 &\checkmark & &25.2$^\ast$ &\underline{88.3} &87.1 &\textbf{89.5} &\underline{84.1}\\
GFA &AAAI'24 &- &512$\times$512 &\checkmark & &- &88.2 &\underline{87.4} &88.9 &82.4\\
GSA-Net &CVPR'24 &MiT-b2 &512$\times$512 &\checkmark & &38.2 &88.2 &\underline{87.4} &\underline{89.0} &82.3\\
DPA &CVPR'24 &VGG-16 &512$\times$512 &\checkmark & &19.5$^\ast$ &87.6 &86.8 &88.4 &83.4\\
\midrule
\textbf{FakeFlow} & &MiT-b2 &512$\times$512 &\checkmark & &29.5$^\ast$ &\textbf{88.5} &\textbf{88.0} &\underline{89.0} &\textbf{84.6}\\
\bottomrule
\end{tabular}
\caption{Quantitative evaluation on the DAVIS 2016 validation set and FBMS test set. OF and PP indicate the use of optical flow estimation model and post-processing technique, respectively. $\ast$ denotes speed calculated on our hardware.}
\label{table1}
\end{table*}

\begin{table*}[t]
\centering 
\small
\begin{tabular}{p{1.5cm}P{1.8cm}P{8mm}P{8mm}P{8mm}P{8mm}P{8mm}P{8mm}P{8mm}P{8mm}P{8mm}P{8mm}|P{8mm}}
\toprule
Method &Backbone &Aero. &Bird &Boat &Car &Cat &Cow &Dog &Horse &Motor. &Train &Mean\\
\midrule
MATNet &ResNet-101 &72.9 &77.5 &66.9 &79.0 &73.7 &67.4 &75.9 &63.2 &62.6 &51.0 &69.0\\
RTNet &ResNet-101 &84.1 &80.2 &70.1 &79.5 &71.8 &70.1 &71.3 &65.1 &\textbf{64.6} &53.3 &71.0\\
AMC-Net &ResNet-101 &78.9 &80.9 &67.4 &82.0 &69.0 &69.6 &75.8 &63.0 &63.4 &57.8 &71.1\\
HFAN$^\dagger$ &MiT-b1 &84.7 &80.0 &72.0 &76.1 &76.0 &71.2 &76.9 &\textbf{71.0} &\underline{64.3} &\underline{61.4} &73.4\\
TMO &ResNet-101 &85.7 &80.0 &70.1 &78.0 &73.6 &70.3 &76.8 &66.2 &58.6 &47.0 &71.5\\
GFA &- &\underline{87.2} &85.5 &\textbf{74.7} &\textbf{82.9} &80.4 &\textbf{72.0} &79.6 &\underline{67.8} &61.3 &55.8 &\underline{74.7}\\
DPA &VGG-16 &\textbf{87.5} &\underline{85.6} &70.1 &77.7 &\underline{81.2} &69.0 &\textbf{81.8} &61.9 &62.1 &51.3 &73.7\\
\midrule
\textbf{FakeFlow} &MiT-b2 &86.3 &\textbf{86.1} &\underline{73.9} &\underline{82.3} &\textbf{81.6} &\underline{71.6} &\underline{80.7} &65.7 &55.4 &\textbf{64.2} &\textbf{75.0}\\
\bottomrule
\end{tabular}
\caption{Quantitative evaluation on the YouTube-Objects dataset. Performance is reported with the $\mathcal{J}$ mean. $\dagger$ indicates the use of test-time augmentation strategy.}
\label{table2}
\end{table*}

\subsection{Implementation Details}

\noindent\textbf{Network architecture.}
Following existing approaches, we design our network with a simple encoder-decoder architecture, similar to TMO~\cite{TMO}. There are two encoders, each dealing with appearance and motion information, respectively. The features from the two encoders are then fused at multiple layers and gradually decoded to the final segmentation mask. Similar to existing methods~\cite{TBD, TMO}, we leverage CBAM~\cite{CBAM} layers after every multi-modal fusion process.

\vspace{1mm}
\noindent\textbf{Network training.}
To provide as comprehensive knowledge as possible to the network, we adopt a two-stage training protocol. In the first stage, we use the YouTube-VOS 2018 training set to pre-train the model. As there are multiple objects in each video sequence, we merge all objects into a single object and define the merged object as the salient object. In the second stage, we fine-tune the network with a combination of the DAVIS 2016 training set and the DUTSv2 dataset. We set the mixing ratio to 3:1 for DUTSv2 and DAVIS 2016, respectively. For network optimization, we adopt a cross-entropy loss and Adam~\cite{adam} optimizer with a learning rate of 1e-5. Network training is implemented with two GeForce RTX TITAN GPUs.

\begin{figure*}[t]
\centering
\includegraphics[width=1\linewidth]{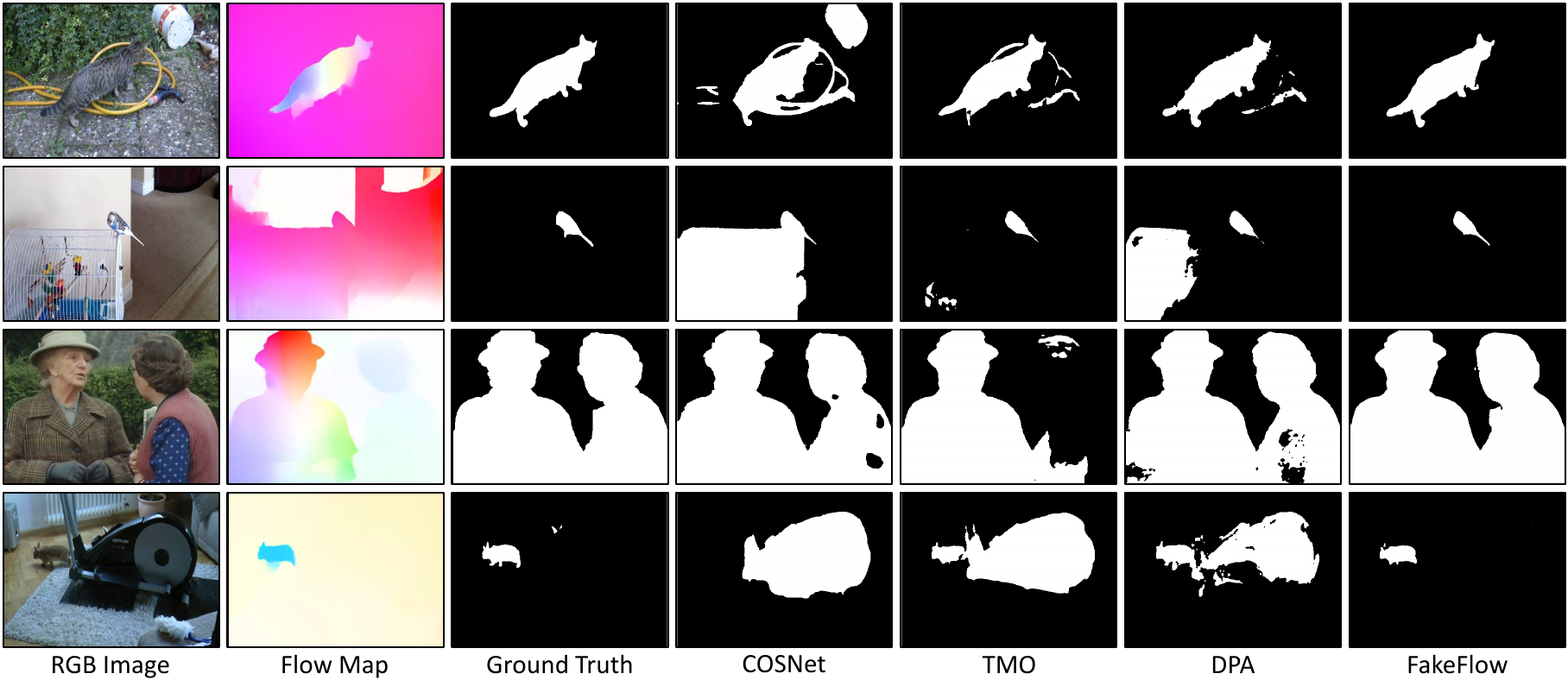}
\caption{Qualitative comparison between state-of-the-art methods and our proposed FakeFlow.}
\label{figure5}
\end{figure*}

\section{Experiments}
We conduct extensive experiments to validate our proposed method, FakeFlow. In this context, D, F, Y, and L represent the DAVIS 2016~\cite{DAVIS} validation set, the FBMS~\cite{FBMS} test set, the YouTube-Objects~\cite{YTOBJ} dataset, and the Long-Videos~\cite{LVID} dataset, respectively. For speed evaluation, we utilize a single GeForce RTX 2080 Ti GPU.

\subsection{Quantitative Results}
In Tables~\ref{table1}, \ref{table2}, and \ref{table3}, we compare our proposed FakeFlow with existing state-of-the-art methods quantitatively. To ensure a fair comparison, we specify the backbone used in each method, including ResNet~\cite{resnet}, DeepLab~\cite{deeplabv3, deeplabv3p}, MiT~\cite{mit}, CSN~\cite{csn}, VGG~\cite{vgg}, and MobileViT~\cite{mobilevit}. The evaluation demonstrates that constructing a robust training data is as crucial as developing a strong architecture. By enhancing only the training data, FakeFlow outperforms all existing methods by a significant margin while maintaining fast inference speeds. Notably, it also achieves segmentation accuracy comparable to that of semi-supervised methods~\cite{agame, stm, cfbi, LVID} on the Long-Videos~\cite{LVID} dataset.

\begin{table}[t!]
\centering 
\small
\begin{tabular}{p{1.8cm}P{2.2cm}P{0.8cm}P{0.8cm}}
\toprule
Method &Backbone &Type &$\mathcal{J}_\mathcal{M}$\\
\midrule
A-GAME &ResNet-101 &SS &50.0\\
STM &ResNet-50 &SS &\underline{79.1}\\
CFBI &DeepLabv3+ &SS &50.9\\
AFB-URR &ResNet-50 &SS &\textbf{82.7}\\
\midrule
AGNN &DeepLabv3 &US &68.3\\
MATNet &ResNet-101 &US &66.4\\
HFAN$^\dagger$ &MiT-b2 &US &\underline{80.2}\\
\midrule
\textbf{FakeFlow} &MiT-b2 &US &\textbf{80.6}\\
\bottomrule
\end{tabular}
\caption{Quantitative evaluation on the Long-Videos dataset. SS and MS denote semi-supervised and unsupervised, respectively. $\dagger$ indicates the use of test-time augmentation.}
\label{table3}
\end{table}

\begin{table}[h]
\centering 
\small
\begin{tabular}{c|c|c|P{5mm}P{5mm}P{5mm}P{5mm}}
\toprule
Version &Pre. &Training &D &F &Y &L\\
\midrule
\Romannum{1} & &VOS &83.2 &66.2 &62.2 &60.5\\
\Romannum{2} & &SOD &79.5 &79.8 &74.9 &74.9\\
\Romannum{3} & &VOS \& SOD &87.6 &82.7 &72.5 &77.2\\
\midrule
\Romannum{4} &\checkmark &- &82.2 &81.7 &73.3 &68.3\\
\Romannum{5} &\checkmark &VOS &87.7 &81.8 &72.8 &73.2\\
\Romannum{6} &\checkmark &SOD &85.8 &83.4 &76.5 &82.0\\
\Romannum{7} &\checkmark &VOS \& SOD &88.5 &84.6 &75.0 &80.6\\
\bottomrule
\end{tabular}
\caption{Ablation study on the training data setting. Pre. indicates the use of pre-training with large-scale video data.}
\label{table4}
\end{table}

\subsection{Qualitative Results}
Figure~\ref{figure5} presents a qualitative comparison between our proposed FakeFlow and existing state-of-the-art solutions. We visualize challenging scenarios where appearance or motion cues are particularly noisy, such as visual distractions or ambiguous motion of the target object. In contrast to other methods that struggle to detect the object accurately under these conditions, FakeFlow consistently produces precise segmentation masks. Notably, our training protocol, which incorporates indistinct optical flow maps due to depth vagueness, enables our model to effectively handle challenging motion cues during inference.

\subsection{Analysis}
\noindent\textbf{Training protocol.} 
In Table~\ref{table4}, we compare various network training protocols. Here, VOS denotes the use of the DAVIS 2016 training set, while SOD refers to the DUTSv2 dataset. The incorporation of generated fake flows significantly enhances model performance, both with and without pre-training on large-scale video data. Notably, the performance improvement on the FBMS test set, YouTube-Objects dataset, and Long-Videos dataset is substantial, given their differing domain properties compared to the DAVIS dataset. Interestingly, the use of our fake image-flow pairs alone surpasses the performance of models trained with the DAVIS dataset on these datasets.

\begin{table}[t]
\centering 
\small
\begin{tabular}{P{1.5cm}|P{6mm}P{6mm}P{6mm}P{6mm}P{6mm}}
\toprule
Backbone &fps &D &F &Y &L\\
\midrule
MiT-b0 &61.1 &86.7 &81.2 &70.4 &75.7\\
MiT-b1 &45.5 &87.3 &81.8 &73.4 &77.4\\
MiT-b2 &29.5 &88.5 &84.6 &75.0 &80.6\\
\bottomrule
\end{tabular}
\caption{Ablation study on the backbone network.}
\label{table5}
\end{table}

\begin{table}[t]
\centering 
\small
\begin{tabular}{p{1.5cm}|P{1.3cm}|P{6mm}P{6mm}P{6mm}P{6mm}}
\toprule
Method &Backbone &fps &$\mathcal{G}_\mathcal{M}$ &$\mathcal{J}_\mathcal{M}$ &$\mathcal{F}_\mathcal{M}$\\
\midrule
\multirow{3}*{HFAN} &MiT-b0 &21.8 &81.2 &81.5 &80.8\\
&MiT-b1 &18.4 &86.7 &86.2 &87.1\\
&MiT-b2 &12.8 &87.5 &86.8 &88.2\\
\midrule
\multirow{3}*{FakeFlow} &MiT-b0 &61.1 &86.7 &86.5 &87.0\\
&MiT-b1 &45.5 &87.3 &87.0 &87.6\\
&MiT-b2 &29.5 &88.5 &88.0 &89.0\\
\bottomrule
\end{tabular}
\caption{Aligned comparison on the DAVIS 2016 validation set. The speed is calculated on the same hardware.}
\label{table6}
\end{table}

\begin{table}[h!]
\centering 
\small
\begin{tabular}{p{1.5cm}|P{6mm}|P{6mm}P{6mm}P{6mm}P{6mm}}
\toprule
Method &Flow &D &F &Y &L\\
\midrule
\multirow{2}*{TMO} & &80.0 &80.0 &73.1 &67.7\\
&\checkmark &86.1 &79.9 &71.5 &72.5\\
\midrule
\multirow{2}*{FakeFlow} & &86.2 &81.2 &75.3 &78.3\\
&\checkmark &88.5 &84.6 &75.0 &80.6\\
\bottomrule
\end{tabular}
\caption{Controlled experiment to evaluate model flexibility.}
\label{table7}
\end{table}

\vspace{1mm}
\noindent\textbf{Backbone network.} We compare various backbone versions in Table~\ref{table5}. Larger backbones demonstrate higher performance across all datasets, although they require more computational time. For our experiments, we utilize the MiT-b2 backbone network as the default.

\vspace{1mm}
\noindent\textbf{Aligned comparison.} To validate the effectiveness of our proposed approach, we present an aligned comparison in Table~\ref{table6}. We select HFAN~\cite{HFAN} as a competitor, as it adopts a two-stream pipeline and per-frame inference protocol similar to our model. Our method consistently outperforms HFAN in both inference speed and segmentation accuracy across all backbone versions. Notably, with a smaller backbone where the potential of pre-trained knowledge is underutilized, HFAN struggles to extract meaningful cues from the limited training data, leading to network overfitting. These results highlight that the design of the training data is crucial, often outweighing the importance of the network architecture itself.

\begin{figure}[t]
\centering
\includegraphics[width=1\linewidth]{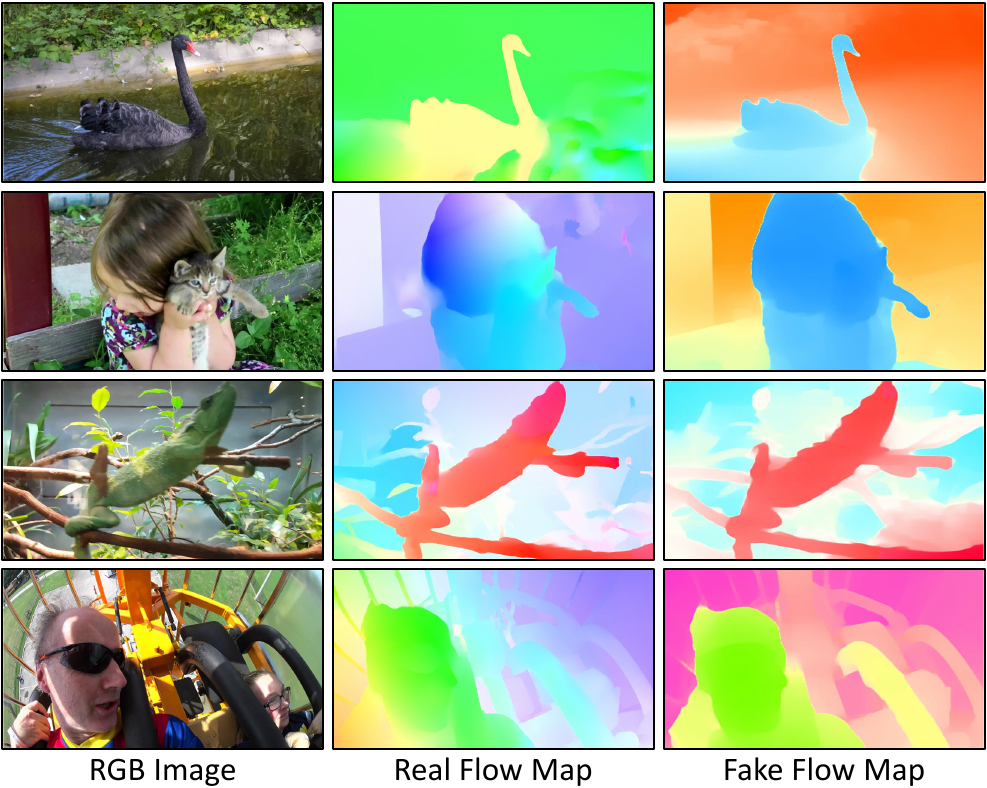}
\caption{Visual comparison between real flow maps from flow estimation model and our generated fake flow maps.}
\label{figure6}
\end{figure}

\vspace{1mm}
\noindent\textbf{Comparison to real flows.} In Figure~\ref{figure6}, we qualitatively compare real optical flows (estimated flows from a pre-trained flow estimation model) with the generated fake flows (simulated using depth-to-flow conversion) on the DAVIS 2017~\cite{DAVIS17} dataset. The figure illustrates that the fake flow maps exhibit highly similar characteristics to the real flow maps.

\vspace{1mm}
\noindent\textbf{Model flexibility.} While our method requires image-flow pairs for mask prediction, we can simulate these pairs by generating fake flows from each video frame. This allows our method to operate without real optical flows, effectively functioning as an image SOD network. To assess the model's flexibility, we compare our method with TMO~\cite{TMO}, which also provides a versatile solution independent of motion availability, as shown in Table~\ref{table7}. Across all benchmark datasets, our method outperforms TMO both with and without optical flow maps, demonstrating its high flexibility. Note that the motion augmentation process for FakeFlow without real flows causes some fluctuations in performance, but these differences are quite negligible.

\begin{figure}[t]
\centering
\includegraphics[width=1\linewidth]{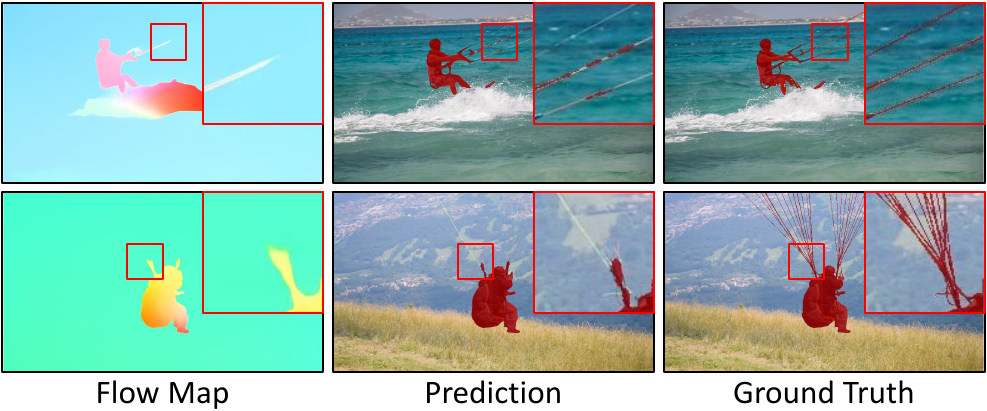}
\caption{Visualized failure cases of our method.}
\label{figure7}
\end{figure}

\vspace{1mm}
\noindent\textbf{Limitation.} Figure~\ref{figure7} visualizes some failure cases of our method. Similar to other two-stream methods leveraging optical flow maps, our approach struggles to capture thin or small parts. These components are often undetected in the optical flows, which can introduce noise during inference.

\section{Conclusion}
In unsupervised VOS, the scarcity of training data has been a significant bottleneck in achieving high segmentation accuracy. Inspired by observations on two-stream approaches, we introduce a novel data generation method based on the depth-to-flow conversion process. Our method outperforms all existing methods on public benchmark datasets without relying on complex modules or heavy post-processing techniques. Our study demonstrates that providing a sufficient amount of high-quality data samples is as important as designing a robust network. We believe that our data generation method can serve as a solid baseline for future research.

\bibliography{aaai25}
\end{document}